\RequirePackage{fix-cm}
\documentclass[smallcondensed]{svjour3}     
\smartqed  

\usepackage{natbib}
\usepackage{hyperref}
\usepackage{graphicx}
\usepackage{enumitem}
\usepackage{color,colortbl}
\usepackage{framed}
\usepackage{makecell}
\usepackage{multirow}
\usepackage[T1]{fontenc}
\usepackage{float}
\usepackage{amsmath,amssymb}
\usepackage[export]{adjustbox}
\usepackage[table,xcdraw]{xcolor}
\usepackage{lscape}
\usepackage{array} 
\usepackage{afterpage}
\usepackage{booktabs}
\usepackage{threeparttable}
\usepackage{xcolor}

\authorrunning{Wong et al.}

\def\makeheadbox{\relax}

\begin{document}

\title{Deep Multiagent Reinforcement Learning: Challenges and Directions}
\titlerunning{Deep multiagent Reinforcement Learning}

\author{Annie Wong \and Thomas Bäck \and Anna V. Kononova \and Aske Plaat
}


\institute{
          Leiden Institute of Advanced
          Computer Science
          \\
          Leiden University, Leiden  \\
          The Netherlands \\
         \\ \email{a.s.w.wong@liacs.leidenuniv.nl}
          }

\date{\today}

\maketitle

\begin{abstract}
This paper surveys the field of deep multiagent reinforcement learning. The combination of deep neural networks with reinforcement learning has gained increased traction in recent years and is slowly shifting the focus from single-agent to multiagent environments. Dealing with multiple agents is inherently more complex as (a) the future rewards depend on multiple players' joint actions and (b) the computational complexity increases. We present the most common multiagent problem representations and their main challenges, and identify five research areas that address one or more of these challenges: centralised training and decentralised execution, opponent modelling, communication, efficient coordination, and reward shaping. We find that many computational studies rely on unrealistic assumptions or are not generalisable to other settings; they struggle to overcome the curse of dimensionality or nonstationarity. Approaches from psychology and sociology capture promising relevant behaviours, such as communication and coordination, to help agents achieve better performance in multiagent settings. We suggest that, for multiagent reinforcement learning to be successful, future research should address these challenges with an interdisciplinary approach to open up new possibilities in multiagent reinforcement learning. 

\end{abstract}

\section{Introduction}
\label{sec:introduction}

Reinforcement learning (RL) is a machine-learning method in which one agent or a group of agents maximises its long-term return through repeated interaction with its environment. Agents are not told what actions to take and must learn its optimal behaviour via trial-and-error. Since rewards may be delayed, an agent has to make a trade-off between exploiting states with the current highest reward and exploring states that may potentially yield higher rewards \citep{bellman1957markovian}. As agents learn by receiving rewards for desirable actions and penalties (negative rewards) for undesired actions, RL can automate learning and decision-making without supervision or having complete models of the environment. However, one drawback of RL methods is that they suffer from the \emph{curse of dimensionality} \citep{bellman1957markovian}: algorithms become less efficient as the dimensions of the state-action space increase \citep{sutton1998introduction}. In recent years the rise of deep reinforcement learning (DRL), a combination of RL and deep learning, has enabled artificial agents to surpass human-level performance in a wide range of complex decision-making tasks, such as in the board game Go \citep{silver2016mastering} and the card game Poker \citep{brown2018superhuman, brown2019superhuman, bowling2015heads}. While prior RL applications required carefully handcrafted features based on human knowledge and experience \citep{sutton1998introduction}, deep neural networks can automatically find low-dimensional representations (features) of high-dimensional data \citep{lecun2015deep}. This development has led to enormous growth in applying RL to more complicated problems. First in single-agent settings such as playing Atari \citep{mnih2015human}, resource management \citep{ wen2015optimal, mao2016resource}, indoor robot navigation \citep{zhu2017target}, cyber security \citep{huang2022reinforcement}, and trade execution \citep{nevmyvaka2006reinforcement}, and more recently in multiagent settings such as bidding optimization \citep{jin2018real}, traffic-light control \citep{chu2020multi}, autonomous driving \citep{sallab2017deep}, financial market trading \citep{bao2019multiagent}, energy usage \citep{prasad2019multi}, fleet optimization \citep{lin2018efficient} and strategy games like Dota 2 \citep{berner2019dota} and Starcraft \citep{vinyals2019grandmaster}. 

It is challenging to translate the successes of DRL in single-agent settings to a multiagent setting. Multiagent reinforcement learning (MARL) differs from single-agent systems foremost in that the environment's dynamics are determined by the joint actions of all agents in the environment, in addition to the uncertainty already inherent in the environment. As the environment becomes nonstationary, each agent faces the moving-target problem: the best policy changes as the other agents' policies change  \citep{busoniu2008comprehensive,papoudakis2019dealing}. The violation of the stationarity assumption required in most single-agent RL algorithms poses a challenge in solving multiagent learning problems. The curse of dimensionality is also worse in a multiagent setting as every additional agent increases the state-action space. At the same time, MARL introduces a new set of opportunities as agents may share knowledge and imitate or directly learn from other learning agents \citep{da2019survey, ilhan2019teaching}, which may accelerate the learning process and subsequently result in more efficient ways of arriving at a goal.

Deep multiagent reinforcement learning (DMARL) constitutes a young field that is rapidly expanding. Many real-world problems can be modelled as a MARL problem, and the emergence of DRL has enabled researchers to move from simple representations to more realistic and complex environments. This survey examines current research areas within DMARL, addresses critical challenges, and presents future research directions. Earlier surveys were driven by the theoretical difficulties
in multiagent systems, including nonstationarity (Hernandez-Leal et al., 2019a; Papoudakis et al., 2019), partial observability, and continuous state and action spaces (Nguyen et al., 2020). Others focus on how agents learn, such as transfer learning (Da Silva and Costa, 2019), modelling other agents (Albrecht and Stone, 2018), or a theoretical domain such as game theory (Yang and Wang, 2021) and evolutionary algorithms \citep{bloembergen2015evolutionary}. A number of studies have looked into the applications of MARL \citep{canese2021multi, feriani2021single, du2021survey}. This paper complements a group of surveys that provides a general framework to classify the deep learning algorithms used in recent DMARL studies \citep{hernandez2019survey, gronauer2021multi}.

When working on this survey, Google Scholar was the leading search engine for finding relevant papers containing keywords such as "multiagent" or "multiagent", "reinforcement learning", and "deep learning". We cover works from leading journals, conference proceedings, relevant arXiv papers, book chapters, and PhD theses. 
We carefully evaluated the studies that came to our attention and developed a taxonomy based on the prominent research directions in the field.

In contrast to prior surveys, we propose a taxonomy based on the challenges inherent in multiagent problem formalisations and their solutions. Modelling a multiagent problem differs from the single-agent setting due to the violation of the stationarity assumption and the difference in learning objectives. Hence, alternative problem formalisations and solutions have been introduced. While other taxonomies also start from multiagent problem representations \citep{yang2021overview, zhang2021multiagent}, these studies only focus on Markov and extensive-form games. Recent MARL research has used additional representations to model multiagent problems, such as the decentralised partially observable Markov game and the partially observable Markov game, which we will also cover in this survey.

The remainder of this paper is organised as follows. In \autoref{sec:singleagent} the preliminaries of single-agent RL are discussed. In \autoref{sec:problemrepresentations} we present the most common DMARL problem frameworks. The taxonomy is introduced in \autoref{sec:taxonomy}. The discussion and recommendations for future research are given in \autoref{sec:discussion}. We end with the conclusion in \autoref{sec:conclusion}.

\section{Single-agent Reinforcement Learning}
\label{sec:singleagent}

\subsection{Markov Decision Process}
\label{sec:Markov Decision Process}
Most RL problems can be framed as a Markov decision process (MDP) \citep{bellman1957markovian}: a model for sequential decision making under uncertainty that defines the interaction between a learning agent and its environment. Formally, it can be defined as a tuple 
$\langle S, A, P, R, \gamma \rangle$ where \textit{S} is the set of states, \textit{A} is the set of actions, \textit{P} is the transition probability function, \textit{R} is the reward function and  $\gamma \in $ [0, 1] is the discount factor for future rewards. 
The learning agent interacts with the environment in discrete time steps. At each time step \textit{t}, the agent is in some state $s_{t} \in S$ and selects an action $a_{t} \in A$. At time step $t_{t+1}$ the agent receives a reward  $r_{t+1} \in R$ and moves into a new state $s_{t+1}$. Specifically, the state transition function is defined as $P(s', r|s, a) = Pr \{ S_{t}=s', R_{t}=r | S_{t-1}=s, A_{t-1}=a \} $ and describes the model dynamics. Each state in an MDP has the Markov property, which means that the future only depends on the current state and not on the history of earlier states and actions. MDPs further assume that the agent has full observability of the states and that the environment is stationary: the transition probabilities and rewards remain constant over time. A setting where the agent does not have full observability of the state is called a partially observable Markov decision process (POMDP) \citep{astrom1965optimal}. 

A policy $\pi$ is a mapping from states to probabilities of selecting each action and can be deterministic or stochastic. The goal of the agent is to learn a policy that maximises its performance and is typically defined as the expected return, computed as the expected discounted sum of rewards, in a trajectory $\tau=(s_{0}, a_{0}, s_{1}, a_{1}, ...)$, a sequence of states and actions in the environment:
\begin{equation}
\mathbb{E}_{\tau}\Big[\sum\limits_{t=0}^{T} \gamma^{t} r_{t} \Big].
\end{equation}

The discount factor $\gamma \in [0,1]$ describes how rewards are valued. A $\gamma$ closer to 0 means that the agent places more value on immediate rewards, while a $\gamma$ closer to 1 indicates that the agent favours future rewards. A policy that maximises the function above is optimal and is denoted as $\pi^{*}$.

Most MDP solving algorithms can be divided into one of three groups: value-based, policy-based, and model-based methods. This distinction is based on the three primary functions to learn in RL \citep{graesser2019foundations}. Hybrid forms of the three primary functions also exist. We present a brief overview of each of the three classes. 

\subsection{Value-based Methods}
\label{sec:Value-based Methods}
Value-based methods learn the value function and derive the optimal policy from the optimal value function. There are two kinds of value functions. The state-value function describes how good it is to be in a state, and it is the expected return from being in state \textit{s} and then following policy $\pi$ and is denoted as:
\begin{equation}
v_{\pi}(s)=\mathbb{E}_{s_{0}=s,\tau \sim \pi}\Big[\sum\limits_{t=0}^{T} \gamma^{t} r_{t} \Big].
\end{equation}

The action-value function or sometimes called the Q-function describes how good it is to perform action \textit{a} in state \textit{s} and is denoted as: 
\begin{equation}
q_{\pi}(s, a)=\mathbb{E}_{s_{0}=s,a_{0}=a,\tau \sim \pi}\Big[\sum\limits_{t=0}^{T} \gamma^{t} r_{t} \Big].
\end{equation}

The optimal policy $\pi^{*}$ maximizes the state-value function such that $v_{\pi_*}(s) > v_{\pi(s)}$  for all $\textit{s} \in S$ and all policies $\pi$. If we have the optimal state-value function, the optimal policy can be extracted by choosing the action that gives the maximum action-value for state \textit{s}. This relationship is given by $\pi^{*}= \max\limits_{\pi} v_{\pi}(s) = \max\limits_{\pi} q_{\pi}(s,a)$.

Deep Q-networks (DQN) \citep{mnih2015human} belong to the value-based methods that have become increasingly popular as studies achieved remarkable results in more complicated environments such as Atari games. Recent developments in RL research show a preference for policy-based strategies, even though value-based methods can capture the underlying structure of the environment \citep{arulkumaran2017deep}.

\subsection{Policy-based and Combined Methods}
\label{sec:Policy-based and Combined Methods}
In contrast to value-based methods, policy-based methods search directly for the optimal policy and the output is represented as a probability distribution over actions. The optimal policy is found by optimising a $\theta$-parameterized policy with respect to the objective via gradient ascent. The policy network weights are updated iteratively so that state-action pairs that result in higher returns are more likely to be selected. The objective is the expected return over all completed trajectories and is defined as follows:
\begin{equation}
J(\theta)=\mathbb{E}_{\tau\sim\pi_\theta}\Big[\sum\limits_{t=0}^{T}\gamma^{t}r_{t}\Big].
\end{equation}

Many policy gradients methods build upon REINFORCE \citep{williams1992simple}, one of the first policy gradient implementations which used Monte Carlo sampling to estimate the policy gradient.

Policy gradient methods perform better in continuous and stochastic environments, learn specific probabilities for each action, and learn the appropriate level of exploration \citep{sutton2018reinforcement}. The main limitation of policy gradient methods is the large variance in the gradient estimators \citep{greensmith2004variance} due to sparse rewards and the fact that only a finite set of states and actions are tried. Policy gradient methods are not very sample-efficient since new estimates of the gradients are learned independently from past estimates \citep{konda2003onactor,peters2008natural}.

Actor-critic methods \citep{konda2003onactor, grondman2012survey, bahdanau2017actor} combine policy-based and value-based methods to address these limitations: Actor-critic methods preserve the desirable convergence properties while maintaining stability during learning. Actor-critic methods consist of an actor that learns a policy and a critic that learns a value function to evaluate the state-action pair. The critic approximates and updates the value function parameters $w$ for either the state-value $v(s;w)$ or the action-value $q(a|s;w)$, and the actor updates the policy parameters $\theta$ for $\pi_{\theta}(a|s)$ in the direction suggested by the critic.

Popular actor-critic methods include Advantage Actor-Critic (A2C) \citep{wu2017scalable}, Asynchronous Advantage Actor-Critic (A3C) \citep{mnih2016asynchronous}, Proximal Policy Optimization (PPO) \citep{schulman2017proximal}, Soft Actor-Critic (SAC) \citep{haarnoja2018soft} and Twin-Delayed Deep Deterministic Policy Gradient (TD3) \citep{dankwa2019twin}. In A3C, multiple agents interact with a copy of the environment in parallel and update the global network parameters asynchronously \citep{mnih2016asynchronous}. In contrast, A2C performs the global network updates synchronously and is found to be more efficient on a GPU machine or when larger policies are trained \citep{openaia2c}. PPO builds upon Trust Region Policy Optimization (TRPO) \citep{schulman2015trust}, a method in which the gradient steps are constrained to prevent destructive policy updates. PPO uses first-order optimisation to compute the updates, simplifying the algorithm's tuning and implementation. In contrast to previous methods, SAC and TD3 are off-policy methods that efficiently reuse past experiences. SAC uses entropy maximization to encourage exploration, while TD3 is a combination of continuous Double Deep Q-Learning \citep{van2016deep}, Policy Gradient \citep{silver2014deterministic} and Actor-Critic \citep{sutton1999policy}.

\subsection{Model-based Methods}
\label{sec:Model-based Methods}
Model-based approaches learn a model of the environment that captures the transition and reward function. The agent can then use planning, the construction of trajectories or experiences using the model \citep{hamrick2021role} to find the optimal policy. While model-free methods focus on learning, where the agent improves a policy or value function from direct experiences generated by the environment, model-based methods focus on planning \citep{sutton2018reinforcement}.

The environment model can either be given or learned. Games such as chess and Go belong to the first category. When there is no given model, the agent must learn it through repeated interaction with the environment using a base policy $\pi_{0}(a_{t}|s_{t})$. The experiences are stored in historical data $\mathcal{D}={(s_t^i, a_t^i,} s_{t+1}^i)$, which is then used to learn the dynamics model $P(s,a)$ by minimizing  $\sum_i ||P(s_t^i, a_t^i)- s_{t+1}^i ||^2$. Given the current state $s$ and action $a$, the next state $s_{t+1}$ is then given by $s_{t+1} = P(s_{t}, a_{t})$. Planning is then performed through $P(s, a)$ \citep{levine2017notes, chua2018deep}. Planning methods generally compute value functions via updates or backup operations to simulated experiences to find the optimal policy \citep{sutton2018reinforcement}.

Examples of model-based algorithms include AlphaZero \citep{silver2017mastering} and MuZero \citep{schrittwieser2020mastering} that achieved state-of-the-art performance in Atari, Go, chess and Shogi. For a recent overview of deep reinforcement learning in model-based games, see \citep{plaat2020learning}.

The main advantage of model-based approaches is better sample efficiency. Agents may use the model to simulate experiences to have fewer interactions with the environment, resulting in faster convergence. However, it is difficult to accurately represent the model, especially in real-world scenarios where the transition dynamics are unavailable. In addition, when bias and inaccuracies are present in the model, errors may accumulate for each step \citep{graesser2019foundations}.

\section{Multiagent Problem Representations} \label{sec:problemrepresentations}
In MARL, a set of autonomous agents interact within the environment to learn how to achieve their objectives. While MDPs have proven helpful in modelling optimal decision-making in single-agent stochastic environments, multiagent environments require a different representation. The state dynamics and expected rewards change upon all agents' joint action, violating the core stationarity assumption of an MDP.

\begin{figure*}[!bt]
\center
\includegraphics[width=\textwidth,trim={4mm 0mm 4mm 0mm},clip]{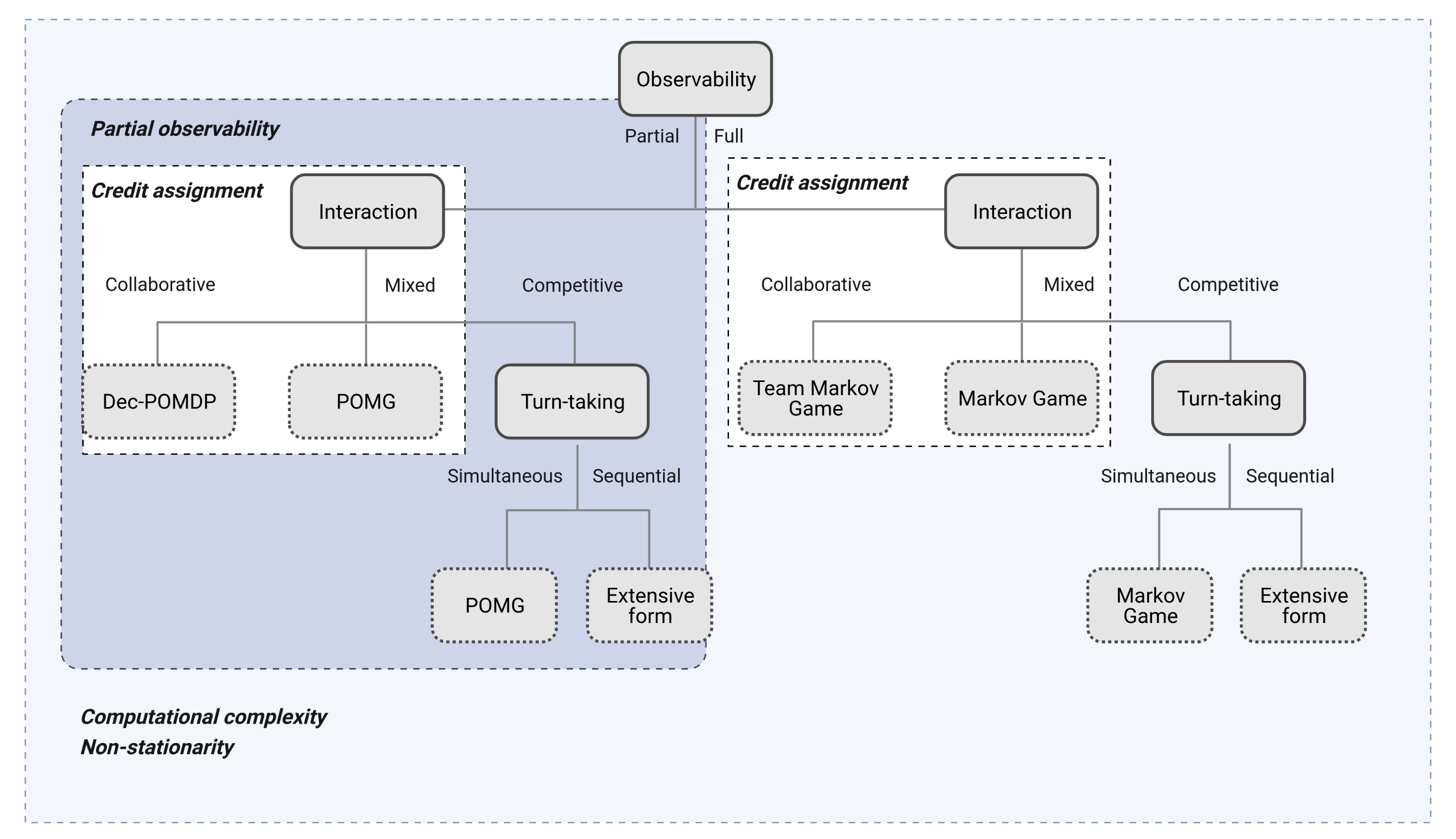}
\caption{\textbf{Diagram of problem representations and their main challenges }
multiagent problem representations can be categorised along a number of axes. First, whether the environment is fully or partially observable. Second, whether the nature of the interaction is collaborative, mixed or competitive. Third, whether turns are taken sequentially or simultaneously. Different problem representations come with different challenges. The four main challenges include computational complexity, nonstationarity, partial observability and credit assignment. Computational complexity and nonstationarity are challenges found in all problem representations, while partial observability and credit assignment are specific to some.\protect\footnotemark[1]} 
\label{fig:overview_paper}
\end{figure*}
\footnotetext[1]{Illustrations are created with BioRender.com}

\begin{figure*}
\center
\includegraphics[width=0.99\textwidth,trim={6mm 0mm 6mm 0mm},clip]{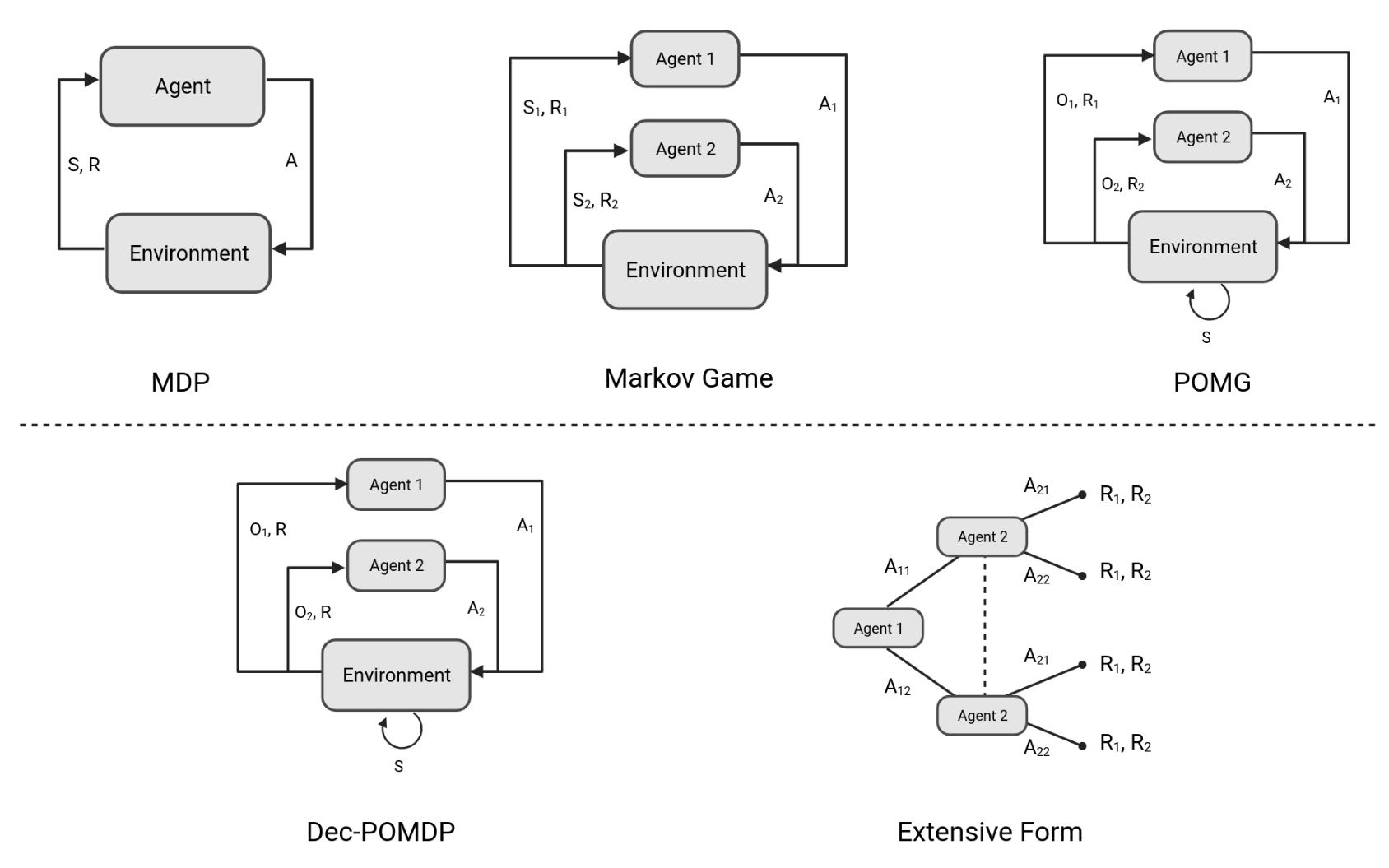}
\caption{
\textbf{Visual depiction of the main problem representations in multiagent reinforcement learning} The MDP is the primary framework used in the single-agent setting. An agent is in some state $S$, performs action $A$, and receives a reward $R$ from the environment. In partially observable environments, the agent cannot view the true state $S$ and receives an observation $O$ instead. For simplicity, all figures display the interaction between two agents $i=1,2$ but can be extended to more agents.}
\label{fig:visualproblemrepresentations}
\end{figure*}

MDPs can be fully or partly visible to the agent. In a multiagent setting, the problem representation is also dependent on the nature of the interaction between agents, which can be cooperative, competitive or mixed, and whether agents take actions sequentially or simultaneously. Figure \ref{fig:overview_paper} shows an overview of the most common theoretical frameworks used in the DMARL literature. When agents have full observability of the state, the problem is usually represented by a Markov game. A particular type is the team Markov game, where agents collaborate to maximise a common reward. If agents collaborate but execute actions decentrally, it is represented by a decentralised partially observable Markov decision process. The partially observable variant for the mixed and competitive setting is known as the partially observable Markov game. The extensive-form game representation is used when agents take turns sequentially instead of simultaneously. The following sections outline the theoretical frameworks pertinent to the DMARL literature, which are visually depicted in Figure \ref{fig:visualproblemrepresentations}. 

\subsection{Markov Games}
Markov games \citep[e.g.][]{littman1994markov}, or Stochastic games \citep{shapley1953stochastic}\footnote[2]{The terms Markov game and stochastic game are used interchangeably in the literature. For consistency, we will continue using the term Markov game throughout the paper.}, provide a theoretical framework to study multiple interacting agents in a fully observable environment and can be applied to cooperative, collaborative and mixed settings. A Markov game is a collection of normal-form games (or matrix games) that the agents
play repeatedly. Each state of the game can be viewed as a matrix representation with the payoffs for each joint action determined by the matrices. 

In its general form, a Markov game is a tuple $\langle I, S, A, R, T \rangle$ where $I$ is the set of $N$ agents, $S$ is a finite state space,  $A = A_{1} \times A_{2} \times ... \times A_{N}$ is the joint action space of $N$ agents, $R = (r_{1}, r_{2}, ..., r_{N})$ where $r_{i}: S \times A \to \mathbb{R}$ is each agent's reward function and $T: S \times A \times S \to [0,1]$ is the transition function. In a team Markov game, agents work together to achieve a goal and share the rewards function $r_{1} = r_{2} = \ldots = r_{N}$. A competitive Markov game is represented by a zero-sum game: the gains for one party automatically result in equal losses for the other. A Markov game is a normal form game, which means that the game is represented in a tabular form, and all agents take their actions simultaneously. 

One way to solve Markov games is to learn equilibria by optimising over an agent's reward function and ignoring others in the environment \citep{tan1993multi,littman1994markov}. Another approach involves best response learners. Agents optimise their reward function while accounting for other agents' changing policies. If these algorithms converge during the play, then it must be an equilibrium \citep{bowling2001rational,bowling2002multiagent}. However, equilibrium concepts either assume infinite computational resources or have been applied to smaller grid-word environments, as they do not scale well with the number of agents. 

The majority of studies in DMARL focus on Markov games, such as Pong \citep{diallo2017learning}, predator games \citep{zheng2018weighted} and the iterated prisoner's dilemma \citep{foerster2018learning}. 

\subsection{Extensive-Form Games}

When agents take turns sequentially, this is modelled as an extensive-form game \citep{kuhn1953contributions}. An extensive-form game specifies the sequential interaction between agents in the form of a game tree. The game tree shows the order of the agents' moves and the possible actions at each point in time. Formally, an extensive-form game with finite and perfect information is given by the tuple $\langle P, A, H, Z, \chi, \rho, \sigma, u \rangle$ where $P$ is a set of players or agents, $A$ is a single set of actions, $H$ is a set of non-terminal choice nodes, $Z$ is a set of terminal outcome nodes, $\chi: H \rightarrow 2^{A}$ is an action function, representing the set of possible actions at each node, $\rho : H \rightarrow P$ is the player function, which assigns at each choice node a player $i \in P$ who is to take action at a given non-terminal node, $\sigma: H \times A \rightarrow H \cup Z$ is the successor function, that maps a choice node and an action to a new choice node or terminal node and $u$ is a set of utility functions \citep{shoham2008multiagent}.

When agents have incomplete information or a partial view of the global state, this can be formalised as an imperfect information extensive-form game in which decision nodes are portioned into information sets. When the game reaches the information set, the agent whose turn it is cannot distinguish between nodes within the information set nor tell which node in the tree has been reached. Formally, an imperfect information extensive-form game is a tuple $\langle P, A, H, Z, \chi, \rho, \sigma, u, I\rangle$ where $ \langle P, A, H, Z, \chi, \rho, \sigma, u \rangle$ is a perfect information extensive-form game and $I = {I_{1}, ..., I_{N}}$ is the set of information partitions of all players.

A strategy maps each agent's information sets to a probability distribution over possible actions. The exploitability is a mean score over all the positions against a worst-case adversary who uses at each turn a best-response. In a Nash equilibrium, the exploitability is equal to 0, and no agents have an incentive to change their strategies \citep{johanson2013evaluating}. 
Studies try to solve extensive-form games by approximating a Nash equilibrium, predominantly in the poker domain \citep{bowling2015heads, heinrich2015fictitious, moravvcik2017deepstack, heinrich2016deep,brown2018superhuman,brown2019superhuman} and board games such as Go \citep{silver2016mastering,silver2017mastering} and Othello \citep{van2013reinforcement}.

\subsection{Decentralized Partially Observable Markov Decision Process}

In a decentralised partially observable Markov decision process (Dec-POMDP), all agents attempt to maximise the joint reward function while having different individual objectives \citep{bernstein2002complexity}. 

A Dec-POMDP is defined by the tuple $\langle I, S, A, \Omega, O, T, R \rangle$, where I is the set of $N$ agents, $S$ is the finite state space, $A$ is the joint action set, $\Omega$ is the joint observations set, $O$ is the observation probability function: $O: \Omega \times A \times S \rightarrow [0, 1]$ and  $O(o_{1},...,o_{N}|a_{1},...,$ $a_{N},s')$ are observed by agents $1, ..., N$, respectively, given that each action tuple $\langle a_{1},...,a_{N} \rangle$ was taken and led to state $s'$. Each agent $i$ has a set of actions $A_{i} \in A$ for each observation $\Omega_{i} \in \Omega$. $T$ is the state transition probability function $T:S\times A \times S \rightarrow [0, 1]$  that specify the transition probabilities $P(s'|s,a_{1},...,a_{N})$. Finally, $R$ is the reward function $R(s,a_{1},...,a_{N})$. 

At every time step, each agent takes an action and receives a local observation that is correlated with the state and an immediate joint reward. A local policy maps local histories of observations to actions, and a joint policy is a tuple of local policies.

The computational complexity of Dec-POMDPs presents a big challenge for researchers. These problems are not solvable with polynomial-time algorithms, and searching directly for an optimal solution in the policy space is intractable \citep{bernstein2002complexity}. One approach is to transform the Dec-POMDP into a simpler model and solve it with planning algorithms \citep{amato2015scalable, ye2017despot}. For instance, using a centralised controller that receives all agents' private information converts the model into a POMDP, and allowing communication that is free of costs and noise reduces it to a multiagent POMDP (MPOMDP) \citep{amato2015scalable,gupta2017cooperative}. Recent solutions also take advantage of the key assumption that planning can be centralised as long as execution is decentralised.

The Dec-POMDP has been used to represent riddles \citep{foerster2016learning}, coordination of bipedal walkers \citep{gupta2017cooperative} and real-time strategy games such as Starcraft \citep{vinyals2019grandmaster, schroeder2019multi,du2019liir}, Dota 2 \citep{berner2019dota}, and Capture the Flag \citep{jaderberg2019human}.

\subsection{Partially Observable Markov Game}
The partially observable Markov game (POMG) \citep{hansen2004dynamic}, also known as the partially observable stochastic game (POSG), is the counterpart of the Dec-POMDP. Instead of a joint reward function, agents optimise their individual reward functions in a partially observable environment. The POMG implicitly models a distribution over other agents' belief states. Formally, a POMG is a tuple $\langle I, S, A, O, {b^{0}}, P, R \rangle$ where $I$ is the set of $N$ agents, $S$ is the set of states, $A_{i}$ is the action set of agent $i$ and $A = A_{1} \times A_{2} \times ... \times A_{N}$ is the joint action set, $O_{i}$ is a set of observations for agent $i$ and $O = O_{1} \times O_{2} \times ... \times O_{N}$ is the joint observation set. The game's initial state, also called the initial belief, is drawn from a probability distribution $b^{0}$ over the states. $P$ is a set of state transitions and observation probabilities, where $P(s', o|s, a)$ is the probability of moving into state $s'$ and joint observation $o$ when taking joint action $a$ in state $s$. $R_{i}: S \times A \rightarrow \mathbb{R} $ is the reward function for agent $i$ where $S$ refers to the joint state $(s_{1},...,s_{N})$ and $A$ refers to the joint actions $(a_{1},...,a_{N})$. The model can be reduced to a POMDP when $|I|=1$.

Dynamic programming algorithms have been developed for POMG \citep{hansen2004dynamic,kumar2009dynamic}, in which agents maintain a belief over the actual state of the environment and other agents' policies. However, applying it to high-dimensional problems becomes intractable, and assumptions are often relaxed or applied to simpler problems. Complexities such as competing goals, nonstationarity and incomplete information make the problem even harder. Examples of POMG include autonomous driving \citep{palanisamy2020multi} and partially observable grid world games \citep{moreno2021neural}. 

\section{Taxonomy of Deep multiagent Reinforcement Learning Algorithms}
\label{sec:taxonomy}
\begin{figure*}[!tb]
\center
\includegraphics[width=\linewidth,trim={10mm 0mm 10mm 0mm},clip]{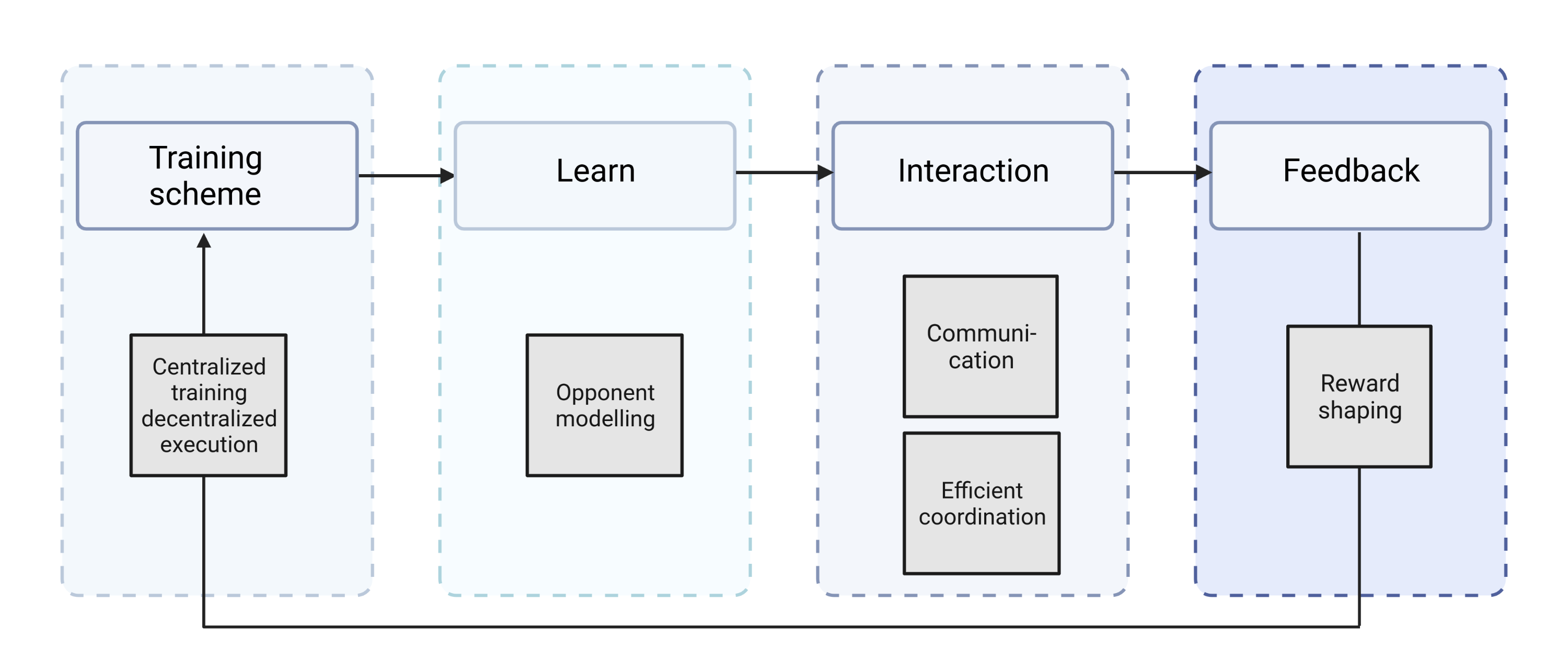}
\caption{
\textbf{Overview of taxonomy}
This figure shows how the paper is organised. We start by discussing the main training scheme in DMARL: centralised training and decentralised execution. We then move to how agents learn through opponent modelling and interact with other agents via communication and coordination. Finally, we discuss how different reward shaping methods act as a feedback mechanism.} 
\label{fig:process}
\end{figure*}

\begin{figure*}
\center
\includegraphics[width=0.95\textwidth,max height=174mm,max width=234mm]{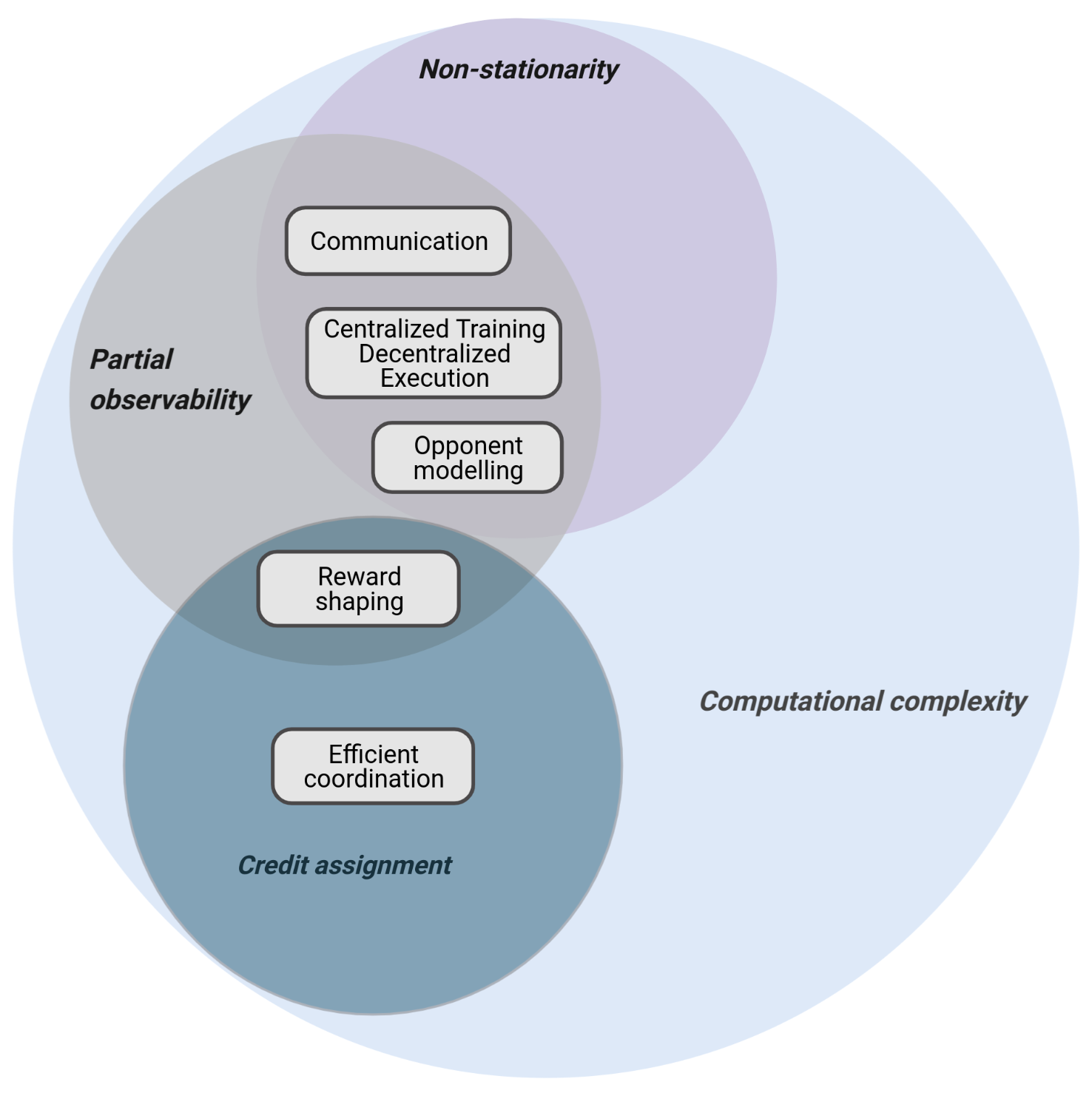}
\caption{
\textbf{Venn diagram of challenges and solutions} The taxonomy of DMARL algorithms comprises five groups: centralised training and decentralised execution, opponent modelling, communication, efficient coordination and reward shaping. Approaches may tackle one or more challenges: nonstationarity, partial observability, credit assignment and computational complexity. Computational complexity is a universal challenge for all approaches. This Venn diagram shows the relations between the surveyed groups of studies and the addressed challenges.}
\label{fig:challenges_solutions}
\end{figure*}

We will now introduce the taxonomy of this paper. We first discuss the four main challenges inherent in multiagent settings: (1) computational complexity, (2) nonstationarity, (3) partial observability and (4) credit assignment. We then provide an overview of current deep learning approaches and discuss how these algorithms address these challenges. The surveyed studies cover the whole learning process of an agent: starting from the training scheme, how it learns and interacts with the environment, to how an agent incorporates feedback, as shown in Figure \ref{fig:process}. The reviewed algorithms have been categorised into one of the following groups: (1) centralised training and decentralised execution, (2) opponent modelling, (3) communication, (4) efficient coordination and (5) reward shaping. Figure \ref{fig:challenges_solutions} shows the relationship between the reviewed studies and the challenges that they address. Finally, Table \ref{table:problems_examples} presents examples of some of the major studies along with their main challenges and solutions.

\afterpage{
\begin{landscape}

\begin{table*}[!bth]
\centering
\caption{Overview of studies along with the problem representation, main challenges, evaluation domains and solutions}
\label{table:problems_examples}
\begin{tabular}{lm{18mm}m{25mm}m{30mm}m{18mm}m{33mm}}
\hline
\multicolumn{1}{c}{\textbf{Study}} & \centering{\textbf{Evaluation domain}} & \centering{\textbf{Problem representation}} & \centering{\textbf{Main challenge(s)}} & \centering{\textbf{Approach}} & \multicolumn{1}{c}{\textbf{Method}}\\
\hline
Moreno et al. 2021 & Running with Scissors & POMG & Partial observability & Opponent modelling & Learning recursive belief models \\
Sukhbaatar, 2016 & Traffic Junction & Dec-POMDP & Partial observability, non-stationarity & Communi-cation & Communication using backpropagation\\
Sunehag et al., 2017 & Switch & Dec-POMDP & Partial observability, non-stationarity & Centralized training and decentralized execution & Value-Decomposition Networks \\
Heinrich and Silver, 2016 & Leduc Poker & Incomplete information extensive-form game & Partial observability, computational complexity & Opponent modelling & Neural Fictious Self-Play \\
Bowling et al., 2015 & Heads-up limit hold'em Poker & Incomplete information extensive-form game & Partial observability, computational complexity & Opponent modelling & Self-play based on counterfactual regret minimization\\
Silver et al., 2018 & Go & Complete information extensive-form game & Computational complexity & Opponent modelling & Self-play and Monte Carlo Tree Search \\
Nguyen et al., 2018 & Matching taxi supply and demand & Dec-POMDP & Credit assignment, partial observability, non-stationarity & Reward shaping & Difference rewards: wonderful life utility and aristocratic utility 
\\
Leibo et al., 2017 & Sequential Social Dilemma & Markov game & Credit assignment & Efficient coordination & Learn policy dynamics of DQN agents by altering variables 
\\
Vinyals et al., 2019 & StarCraft & \multirow{2}{*}{Dec-POMDP} & Credit assignment, partial observability, & \multirow{2}{18mm}{Opponent modelling} & \multirow{2}{*}{Population-based self-play} \\
Jaderberg et al., 2019 & Capture the flag &  & nonstationarity, computational complexity  \\

Berner et al., 2019 & Dota 2 & Dec-POMDP & Credit assignment, partial observability, nonstationarity, computational complexity & Opponent modelling & OpenAI Five: self-play against itself and past selves \\
\hline
\end{tabular}%
\end{table*}
\end{landscape}
}

\subsection{Challenges}
Reinforcement learning in a multiagent environment comes with numerous challenges. Addressing these challenges is a prerequisite for the development of effective learning approaches. Despite promising results in the literature, computational complexity, nonstationarity, partial observability, and credit assignment remain largely unsolved. 

The four challenges do not occur in isolation. In contrast, a multiagent problem usually deals with one or more challenges simultaneously. All multiagent problems deal with high computational demands, and the higher the number of agents, the more demanding it is on computing power. The problem of nonstationarity can lead to an infinite loop of agents adapting to other agents  \citep{papoudakis2019dealing}, and this problem is exacerbated when agents have only a partial view of the state, which means they have less information, and it is even harder to distinguish the effects of their actions from that of other agents. Consequently, agents cannot distil the individual contribution to the team reward, also known as the credit assignment problem. We turn to each of these aspects next. 

\subsubsection{Computational Complexity}
A current limitation of RL algorithms is the low sample efficiency, which requires an agent to interact a vast amount of times with the environment to learn a useful policy \citep{yu2018towards}. For example, to teach an agent to play the game of Pong, at least ten thousand samples are needed, while humans, on average, can master the game in dozens of trials \citep{ding2020challenges}. The sample complexity of reinforcement, or the amount of data an agent needs to collect to learn a successful policy \citep{kakade2003sample}, worsens when multiple interacting agents are learning simultaneously. Computational complexity in reinforcement learning is then how much computation, in terms of time and memory requirements, is required to collect sufficient data samples to output an approximation to the target \citep{kakade2003sample}. A challenge of MARL research is to develop algorithms that can handle this high level of computational complexity. In particular, on complex or continuous-space problems, we face slow learning of new tasks and, in the worst-case, tasks even become infeasible to master. Hence, many studies focus on designing better sample efficiency and scalability of algorithms to deal with the computational complexity in reinforcement learning. 

\subsubsection{Nonstationarity}
In a multiagent environment, all agents learn and interact with the environment concurrently. The state transitions and rewards are no longer stationary for an agent since the new state of the environment is dependent on the joint action of all agents instead of the agent's own behavior. Consequently, agents need to keep adapting to other agents' changing policies. The Markov assumption is violated as the state of the environment no longer gives sufficient information for optimal decision-making \citep{van2012reinforcement}, which is problematic since most RL algorithms assume a stationary environment to guarantee convergence.

Recent works have addressed nonstationarity differently, focusing on various variables: such as the setting, which can be cooperative \citep{son2019qtran}, competitive \citep{berner2019dota} or mixed \citep{leibo2017multi}, whether and how opponents are modelled \citep{brown1951iterative, bowling2015heads}, the availability of opponent information \citep{foerster2018counterfactual, he2016opponent}, and whether the execution of actions is centralised \citep{foerster2018counterfactual, lowe2019pitfalls} or decentralised \citep{tan1993multi}. There is also a wide range of sophistication across algorithms: some algorithms ignore that the environment is nonstationary, assuming that other agents are part of the environment, while more complex methods involve opponent modelling with recursive reasoning \citep{hernandez2019survey}. One way to address nonstationarity is to learn as much as possible about the environment, for example, using centralised training with decentralised execution (section \ref{ctde_}), through opponent modelling (section \ref{opponentmodelling}), and exchanging information between agents (section \ref{communication}). For a thorough overview of how algorithms model and cope with nonstationarity, we refer to recent surveys on nonstationarity \citep{papoudakis2019dealing, hernandez2019survey}. 

\subsubsection{Partial Observabilty}
In a partially observable environment, agents cannot access the global state and must make decisions based on local observations. This results in incomplete and asymmetric information across agents, which makes training difficult. Other agents' rewards and actions are not always visible, making it difficult to attribute a change in the environment to an agent's own action. Partial observability has been mainly studied in the setting where a group of agents maximises a team reward via a joint policy (e.g. in the Dec-POMDP setting). The two main approaches for dealing with partial observability are the centralised training and decentralised execution paradigm \citep{kraemer2016multi, mahajan2019maven, foerster2018counterfactual, lowe2017multiagent} and using communication to exchange information about the environment \citep{foerster2016learning, mao2017accnet, peng2017multiagent}.

\subsubsection{Credit Assignment}
Two credit assignment problems are inherent in multiagent settings. The first problem is that an agent cannot always determine its individual contribution to the joint reward signal due to other concurrently acting agents in the same environment \citep{minsky1961steps}. This makes learning a good policy more difficult as the agent cannot tell whether changes in the global reward were due to its own actions or others in the environment. An alternative to the global reward structure is to let agents learn based on a local reward: a reward based on the part of the environment that an agent can directly observe. However, while an agent may increase its local reward more quickly, this approach encourages selfish behaviour that may lower overall group performance. Hence, reward shaping methods, the practice of supplying the agent with additional rewards beyond those given by the underlying environment to improve learning, have been introduced to deal with the credit assignment problem \citep{ng1999policy}.

The second problem involves constructing a reward function to promote effective collaborative behaviour. This is especially difficult when mixed incentives exist in an environment, such as social dilemmas. The lazy agent problem is also undesirable \citep{sunehag2017value}: when multiple agents interact simultaneously, and one agent learns a good policy, the second agent can hold back to avoid affecting the performance of the first agent.

\subsection{Centralised Training and Decentralised Execution}
\label{ctde_}
\begin{figure*}
\center
\includegraphics[width=\linewidth,trim={7mm 0mm 9mm 0mm},clip]{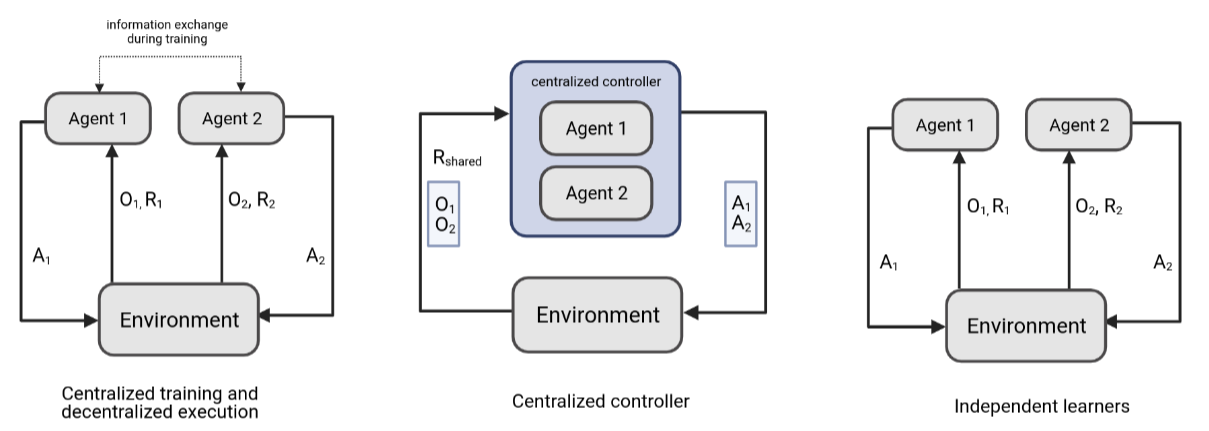}
\caption{
\textbf{Overview of training schemes}
The three main training schemes in multiagent settings are centralised training and decentralised execution, using a centralised controller and independent learning. The most popular approach is centralised training with decentralised execution, where agents can share information during training, but actions are executed decentrally based on local observations.} Using a centralised controller reduces the problem to a single-agent problem but is computationally infeasible. Finally, independent learners consider other agents part of the environment but ignore the nonstationarity problem. 
\label{fig:ctde}
\end{figure*}
We will now turn to the approaches developed to address these challenges.

The main challenge in DMARL is to design a multiagent training scheme that is efficient, and that can deal with the nonstationarity and partial observability problem. Figure \ref{fig:ctde} shows the three most common training schemes. One of the most simple multiagent training schemes is to train multiple collaborating agents with a centralised controller and to reduce it to a single-agent problem. All agents send their observations and policies to a central controller, and the central controller decides which action to take for each agent. This method mitigates the problem of partial observability when agents have incomplete information about the environment. However, using a centralised controller is computationally expensive in large environments and risky as it is a single point of failure. Conversely, all agents can learn an individual action-value function and view other agents as part of the environment \citep{tan1993multi}. This method does not allow agents to coordinate with each other and ignores the nonstationarity problem. 

An approach combining centralised and decentralised processing is centralised training and decentralised execution \citep{kraemer2016multi}. The main idea is that agents can access extra information during training, such as other agents' observations, rewards, gradients and parameters. Agents then execute their policy based on local observations. Centralised training and decentralised execution mitigate nonstationarity and partial observability, as access to additional information during training stabilises agents' learning, even when other agents' policies are changing. Centralised training and decentralised execution methods can be divided into value-based and policy-based methods. Single-agent value-based methods focus on learning and derive the optimal policy via the learned value function. In MARL, cooperating agents have to optimise a team value function, and studies investigate the best way to decompose and optimise this value function. On the other hand, traditional policy-based methods search directly for the optimal policy. In a multiagent setting, nonstationarity makes learning more challenging as all agents update their policies simultaneously. Hence, most policy-based methods use the actor-critic architecture, in which a centralised critic is used to exchange extra information during training.

Value-based methods focus on how to decouple centrally learned value functions and use them for decentralised execution. Value-function factorisation is one of the most popular methods in this category \citep{sunehag2017value,rashid2020monotonic,son2019qtran,mahajan2019maven,rashid2020weighted,yang2020q}. Value Decomposition Networks (VDN) \citep{sunehag2017value} decompose the team value function into a sum of linear, individual value functions. The optimal policy arises by acting greedily with respect to the Q-value, an estimate of how good it is to take an action in a particular state during execution. QMIX \citep{rashid2020monotonic} improves VDN's performance by treating the joint value function as a nonlinear combination of individual value functions and a monotonic constraint. However, this constraint limits the performance of collaborating agents that require significant coordination \citep{rashid2020weighted}. QTRAN \citep{son2019qtran} employs a different factorisation method that can escape the monotonicity and additivity constraints. However, it relies on regularisations to maintain tractable computations, which may impede performance on complex multiagent settings \citep{mahajan2019maven}. Numerous algorithms build further upon QMIX. For instance, Weighted QMIX extends QMIX to nonmonotonic environments by placing more weights on joint actions with higher rewards \citep{rashid2020weighted}. Multiagent Variational Exploration (MAVEN) \citep{mahajan2019maven} addresses the inefficient exploration problem in QMIX via committed exploration: coordinated exploratory actions over extended time steps in dealing with environments that require long-term coordination. MAVEN uses a hybrid value and policy-based method approach by conditioning value-based agents on the shared latent variable controlled by a hierarchical policy. Value-Decomposition Actor-Critic (VDAC) enforces the same monotonic relationship between the global state-value and the local state-values as QMIX. However, unlike QMIX, VDAC is compatible with A2C, which makes sampling more efficient. In addition, the study demonstrates that following a simple gradient calculated from a temporal-difference advantage, the policy can converge to a local optimal \citep{su2021value}. Q-DPP \citep{yang2020multi} do not rely on constraints to decompose the global value function. Instead, it builds upon determinantal point processes: probabilistic models that capture both quality and diversity when a subset is sampled from a ground set, allowing for a natural factorisation of the global value function.

Policy-based methods mainly focus on the actor-critic architecture (see \autoref{sec:Policy-based and Combined Methods}). These studies use a centralised critic to train decentralised actors. Counterfactual multiagent (COMA) \citep{foerster2018counterfactual} uses a centralised critic to approximate the Q-function and decentralised actors to optimise policies. The centralised critic has access to the joint action and all available state information, while each agent's policy only depends on its historical action-observation sequence. Along the same line, multiagent Deep Deterministic Policy Gradient (MADDPG) extends the Actor-Critic algorithm so that the critic has access to extra information during training and the actor only has access to local information \citep{lowe2017multiagent}. As opposed to COMA, which uses one centralised critic for all agents, MADDPG has a centralised critic for each agent to have different reward functions in competitive environments. MADDPG can learn continuous policies, whereas COMA focuses on discrete policies. Several studies build upon MADDPG. For instance, R-MADDPG \citep{wang2020r} extends the MADDPG algorithm to the partially observable environment by having both a recurrent actor and critic that keep a history of previous observations, and M3DDPG \citep{li2019robust} incorporates minimax optimisation to learn robust policies against agents with changing strategies. Since these methods concatenate all the observations in the critic, the input dimension increases exponentially with each agent. Hence, several studies have devised more efficient methods to deal with this problem. For instance, Mean-Field Actor-Critic \citep{yang2018mean} factorises the Q-function using only the interaction with the neighbouring agents based on mean-field theory \citep{stanley1971phase}, and the idea of dropout\footnote[3]{Randomly dropping units in the neural network to avoid overfitting \citep{srivastava2014dropout}.} can be extended to MADDPG to handle the large input space \citep{kim2019message}. 

Centralised training and decentralised execution have been applied to solve complex strategy games such as StarCraft Micromanagement \citep{foerster2018counterfactual} and hide-and-seek \citep{baker2019emergent}.

\subsection{Opponent Modelling}
\label{opponentmodelling}

Opponent modelling belongs to the class of model-based methods \citep{markovitch2005learning} and refers to the construction of models of the beliefs, behaviours, and goals of other agents in the environment \citep{albrecht2018autonomous}. An agent can use these opponent models to guide decision-making. Opponent modelling algorithms generally take a sequence of interactions with the modelled opponent as input and predict action probabilities as output. After generating the opponent's model, an agent can derive its policy based on that model. This method helps an agent discover the competitor's intentions and weaknesses. Learning the model is generally considered more data-efficient than model-free approaches in which the policy is updated from direct observations \citep{markovitch2005learning}. Opponent modelling mitigates the nonstationarity and partially observability problem as agents collect historical observations to learn about the environment (i.e. opponents), allowing agents to track and switch between policies. This method is especially beneficial in the adversarial setting when the opponent has opposing interests, and other approaches such as communication and centralised training that require the opponents' information are unlikely. For a comprehensive overview of opponent modelling, we refer to other work \citep{albrecht2018autonomous}. 

Early opponent modelling methods assumed fixed play of opponents. Neural Fictitious Self-Play (NFSP) extends the idea of fictitious play \citep{brown1951iterative} with neural networks to approach a Nash equilibrium in imperfect information games such as Poker \citep{heinrich2016deep}. The main idea is to keep track of the opponents' historical behaviours and to choose a best response to the opponents' average strategies.

While NFSP requires actual interaction with the opponent, other methods do not. For instance, counterfactual regret minimisation has achieved success in poker \citep{bowling2015heads}. AlphaZero achieved remarkable results in Go, chess, and Shogi, using a neural network with self-play and Monte Carlo Tree Search \citep{silver2017mastering}. MuZero was able to achieve this without a given model. Instead of modelling the entire environment, it focused on the three core elements most relevant for planning: the value, policy and reward \citep{schrittwieser2020mastering}. Still, these studies assume that the opponent follows a stationary strategy. 

Later approaches look at nonstationary environments in which an agent has to track, switch, and possibly predict behaviour. Several studies achieved superhuman performance using self-play in real-time strategy games characterised by long time horizons, nonstationary environments, partially-observed states, and high dimensional state and action spaces. OpenAI Five employs a similar method to fictitious play in playing Dota 2, a video game in which two teams compete to conquer each other's base, but the algorithm learns a distribution over opponents and uses the latest policy instead of the average policy \citep{berner2019dota}. This infrastructure has also been used to solve hide-and-seek, but hide-and-seek agents can act independently as the training scheme is centralised training and decentralised execution \citep{baker2019emergent}. In Capture-the-Flag and StarCraft II, a population of agents is trained to introduce variation. Policies are made more robust by letting agents play with sampled opponents and teammates from this population in a league \citep{jaderberg2019human, vinyals2019grandmaster}. 

Some studies assume that the opponent switches between a set of stationary policies over time \citep{he2016opponent,everett2018learning, zheng2018deep}. These algorithms derive the optimal policy based on the learned opponent's model and identify when the opponent changes the behaviour, and the agent has to relearn a new policy. Over time, the agent has a library of inferred opponent strategies and associated best response policies. The two main challenges are designing a policy detection mechanism and learning a best-response policy. Some studies use a variant of Bayes' rule to learn opponent models and
assign probabilities to the opponent's available actions. An agent starts with a prior belief that is continually updated during interaction to make it more accurate. Switching Agent Model (SAM) learns opponent models from observed state-action trajectories in combination with a Bayesian neural network \citep{everett2018learning}. A Deep Deterministic Policy Gradient algorithm \citep{lillicrap2016continuous} is used to learn the best response. Distilled Policy Network-Bayesian Policy Reuse+ (DPN-BPR+) \citep{zheng2018deep} extends the Bayesian Policy Reuse+ algorithm (BPR+) \citep{hernandez2016bayesian}  with a neural network to detect the opponent's policy via both its behaviour and the reward signal. The latter uses policy distillation \citep{rusu2016policy} to learn and reuse policies efficiently. 
Others use a form of deep Q-learning \citep{mnih2013playing}. Deep Reinforcement Opponent Network (DRON) \citep{he2016opponent} uses one network to learn the Q-values to derive an optimal policy and a second network to learn the opponent policy representation, in addition to expert networks that capture different types of opponent strategies. A drawback of DRON is that it relies on handcrafted opponent features. Previous methods assume that the opponent remains stationary within an episode. Deep Policy Inference Q-Network (DPIQN) and Deep Recurrent Policy Inference Q-Network (DRPIQN) \citep{hong2018deep} incorporate policy features as a hidden vector into the deep Q-network to adapt itself to unfamiliar opponents. DRPIQN uses a Long Short Term Memory (LSTM) layer so agents can learn in partially observable environments. This LSTM layer utilises a recurrent neural network architecture that can take observations as input and allow agents to model time dependencies and capture the underlying state \citep{hausknecht2015deep}.

Previous approaches do not consider an intellectual and reasoning opponent. According to the theory of mind, people attribute mental states to others, such as beliefs, intents and emotions \citep{premack1978does}. These models help to analyse and infer others' behaviours and are essential in social interaction \citep{frith2005theory}. 
Learning with Opponent-Learning Awareness (LOLA) \citep{foerster2018learning} anticipates and shapes opponents' behaviour. Specifically, it includes a term that considers the impact of an agent's policy on the learning behaviour of opponents. One drawback is that LOLA assumes access to the opponent's parameters, which is unlikely in an adversarial setting. 
Others focus on recursive reasoning by learning models over the belief states of other players, a nesting of beliefs that can be represented in the form: "I believe that you believe that I believe" \citep{wen2019probabilistic,tian2021learning}. The Probabilistic Recursive Reasoning (PR2) framework \citep{wen2019probabilistic} first reflects on the opponent's perspective: what the opponents would do given that the opponents know the agent's current state and action. Given the potential actions of the opponent, the agent selects a best response. The recursive reasoning process can be viewed as a hierarchical process with k-levels of reasoning. At level $k=0$, agents take random actions \citep{dai2020r2} or act based on historical interactions, the main assumption in traditional opponent modelling methods \citep{wen2019probabilistic}. At $k=1$, an agent selects its best response to the agents acting at lower levels. Studies show that it pays off to reason about an opponent's intelligence levels \citep{tian2021learning} and that reasoning at a higher level is beneficial as it leads to faster convergence \citep{dai2020r2} and better performance \citep{moreno2021neural}.

\subsection{Communication}
\label{communication}
\begin{figure*}
\center
\includegraphics[width=\linewidth,trim={5mm 7mm 5mm 0mm},clip]{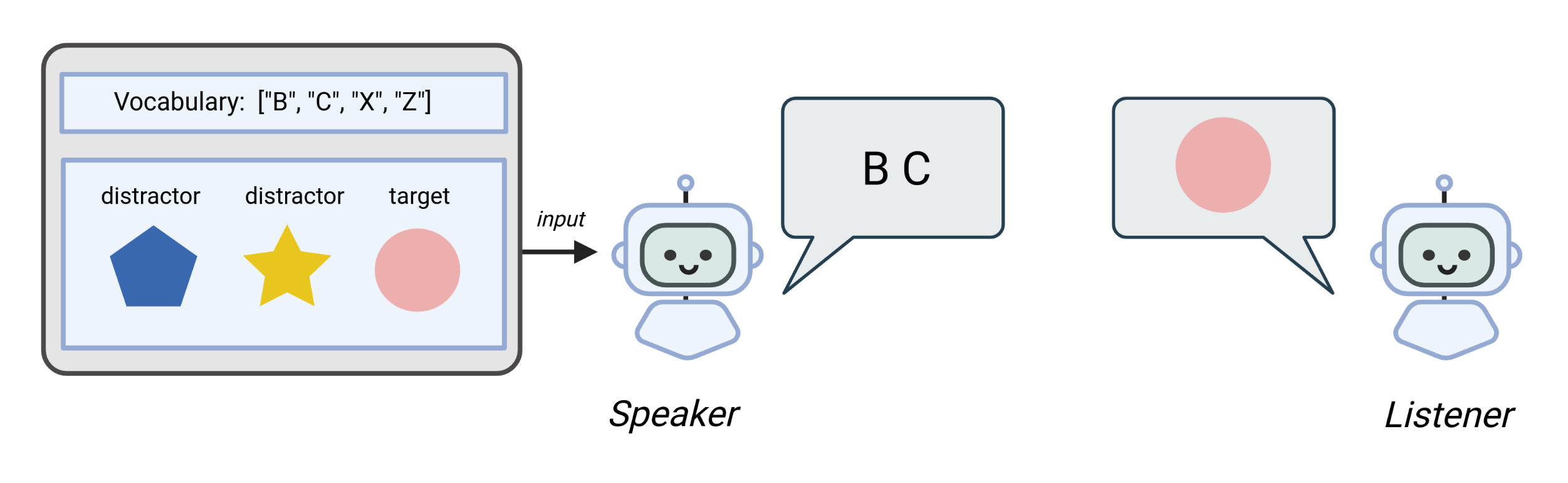}
\caption{
\textbf{Basic referential game}
In this basic referential game example, two agents have to develop a communication protocol so that the speaker can translate the target into a message and the listener can understand which one is the target. The game works as follows. The speaker receives as input three images. One is the target, and the other two are distractions. The speaker has to use the symbols in the vocabulary, which consists of the symbols "B", "C", "X" and "Z", to send a message to the listener. The listener sees the messages and has to guess the target message. If the target is correct, both agents receive a reward.}  
\label{fig:referentialgame}
\end{figure*}
Through communication, agents can pass information to reduce the complexity of finding good policies. For instance, agents exploring different parts of the environment can share observations to mitigate partial observability and share their intents to anticipate each others' actions to deal with nonstationarity. Communication can also be used for transfer learning so that more experienced agents can share their knowledge to accelerate the learning of inexperienced agents \citep{taylor2009transfer}. One of the fundamental questions in communication is how language emerges between agents with no predefined communication protocol \citep{lazaridou2017multiagent}, and, subsequently, how meaning and syntax evolve through interaction \citep{jaques2019social}. Learning this process will help researchers better understand human language evolution and contribute to more efficient problem-solving in a team of interacting agents \citep{lazaridou2020emergent}.

Several studies investigate how agents learn a successful communication protocol. A communication protocol should inform agents which concepts to communicate and how to translate these concepts into messages \citep{hausknecht2016grounded}. Many studies approach this problem as a referential game \citep{lazaridou2017multiagent,havrylov2017emergence}. A referential game involves two or more agents in which speakers and listeners must develop a communication protocol to refer to an object (Figure~\ref{fig:referentialgame}). In the basic version of this game with two agents, the speaker sends two images and a message from a fixed vocabulary to the receiver. One of the images is the target, which the listener has to identify based on the message. Both agents receive a reward when the classification is correct 
\citep{lazaridou2017multiagent}. To succeed in this game, agents must understand the image content and express the content through a common language. The language can be discrete, where messages are a single symbol \citep{lazaridou2017multiagent} or a sequence of symbols \citep{havrylov2017emergence}, or continuous, where messages are continuous vectors \citep{sukhbaatar2016learning}. 

Using DRL, end-to-end policies can be learned in which agents receive image pixels as input and a corresponding message as output. For example, two agents represented as simple feed-forward networks can learn a communication protocol to solve the basic referential game \citep{lazaridou2017multiagent}. Language also emerges in more complicated versions of the game that require dialogue \citep{jorge2017learning, das2017learning,kottur2017natural} or negotiation \citep{cao2018emergent} between agents. Agents trained with deep recurrent Q-networks (DRQN) \citep{jorge2017learning} and REINFORCE \citep{das2017learning,  kottur2017natural} are able to learn a communication protocol from scratch. Since communication is not always meaningful, it is important to develop metrics for emergent communication.
An example is when an agent sends a message that has no actual impact on the environment. Agents with the capacity to communicate should exhibit positive signalling and positive listening \citep{lowe2019pitfalls}. Positive signalling means that messages correlate with observations or actions, and positive listening refers to updating beliefs or behaviour after receiving a message. Most studies focus solely on positive signalling metrics. However, positive signalling may occur without positive listening \citep{lowe2019pitfalls}, which indicates that there was no actual communication. 

In contrast to earlier works that consider communication as the primary learning goal, other works consider communication an instrument to learn a specific task. Most of these studies focus on coordination in collaborative environments and show that communication improves overall performance. Differentiable Interagent Learning (DIAL) \citep{foerster2016learning} uses centralised training and decentralised execution. Communication is continuous during training and discrete during the execution of the task. Continuous communication during training is particularly effective as it enables the exchange of gradients between agents, which improves performance. CommNet shows that the exchange of discrete symbols is less efficient than continuous communication, as the latter enables the use of backpropagation to train agents efficiently \citep{sukhbaatar2016learning}. While DIAL and CommNet base their approach on DQRN,  later studies propose the Actor-Critic architecture, including Actor-Coordinator-Critic Net (ACCNet) \citep{mao2017accnet}, Bidirectionally Coordinated Network (BiCNet) \citep{peng2017multiagent} and MADDPG \citep{lowe2017multiagent}. This architecture can solve more complex problems than previous approaches and works for continuous actions. In addition, when critics are individually learned \citep{jiang2018learning} instead of centrally computed \citep{iqbal2019actor}, agents have different reward functions, which is suitable for competitive settings. 

Communication also allows peer-to-peer teaching. More experienced agents communicate their knowledge to learning agents, accelerating the learning of a new task \citep{da2017simultaneously, omidshafiei2019learning, ilhan2019teaching, amir2016interactive}. However, having agents send messages to all agents is costly and inefficient. Thus, an important question is how to filter the most important messages and to whom to send them. One approach is to limit communication bandwidth \citep{foerster2016learning,kim2020learning} or use a communication budget \citep{ilhan2019teaching, omidshafiei2019learning}. Others use metrics to identify relevant messages, such as attention mechanisms. In its simplest form, this is a vector of importance weights \citep{peng2018learning, gu2021attention, mao2020learning}. An alternative is to keep confidence scores about states \citep{da2017simultaneously}. However, communication comes at a cost and increased complexity. Negative transfer can also happen, for example, when the message contains inaccurate or noisy information so that performance may become worse \citep{taylor2009transfer}. Therefore, it is essential to trade off benefits and costs or find a better way to filter valuable information. 

\subsection{Efficient Coordination}

Another group of studies investigate agents' emergent behaviours and look at how cooperating agents can coordinate actions most efficiently. These studies are conducted in mixed environments with elements of both cooperation and competition. 

A key question is how to design reward functions so that agents adapt to each others' actions, avoid conflicting behaviour and achieve efficient coordination. By engineering the reward function, competitive or cooperative behaviours can be stimulated \citep{tampuu2017multiagent}. While early studies look at how agents can maximise external rewards, recent works assume that agents are intrinsically motivated. 

Most studies look at multiagent behaviour in social dilemmas \citep{eccles2019learning, lerer2018maintaining, leibo2017multi, mckee2020social, jaques2019social,peysakhovich2017prosocial}. Earlier studies, mainly influenced by game theory, have looked at social dilemmas as a matrix game in which agents choose pure cooperation or pure defect. Recent studies generalise these social dilemmas to temporally and spatially extended Markov games, also known as a sequential social dilemma \citep{leibo2017multi}. This setting is more realistic as people can adapt and change their strategies. One notable example is the repeated prisoner's dilemma. In each turn, each agent decides whether to cooperate or defect. When both agents cooperate, both agents get good rewards. Contrary, defection improves one agent's reward at the expense of the other agent. Thus, an agent can decide to retaliate or trust the opponent, dependent on the actions in the previous round. 

One of the first sequential social dilemma studies examined how policies change due to environmental factors or agent properties \citep{leibo2017multi}. They found that agents learn more aggressive policies when resources are limited. In addition, manipulating the discount rate over the rewards, batch size, and the number of hidden units in the network affected emerging social behaviour. While this study took a descriptive approach to understand how behaviours change to different rules and conditions, others took a prescriptive approach in which agents learn to cooperate without being exploited \citep{lerer2018maintaining, wang2018towards}. The general approach comprises two steps: first, detect the level of cooperation of the opponent, and then mimic or reciprocate with a slightly higher-level cooperation policy to induce cooperation without getting exploited. This approach is based on the Tit-for-Tat principle \citep{axelrod1981evolution}: the strategy suggests cooperation in the first round and copies the opponent's behaviour afterwards. 

Previous approaches assume that the only incentive for cooperation is the external reward. However, there is a rapidly growing literature where cooperation occurs from social behaviour and intrinsic motivation \citep{mckee2020social, jaques2019social, peysakhovich2017prosocial,hughes2018inequity}. 

Psychology research has shown that people do not always seek to maximise utility \citep{dovidio1984helping}. In addition, an intrinsic reward may be a good alternative in sparse environments. Several attempts have been made to design these internal rewards. For instance, inequity aversion, which refers to the preference for fairness and resistance against inequitable outcomes \citep{fehr1999theory}, has improved coordination in social dilemmas \citep{hughes2018inequity}. The main idea is to punish agents that deviate too much from the average behaviour. Underperforming and overperforming agents are both undesirable, as the first may exhibit free-riding behaviour while the latter may be operating a defective policy. Another approach is to make agents care about the rewards of teammates \citep{peysakhovich2017prosocial, jaques2019social}. 

Pro-social behaviour improves the convergence probabilities of policy gradient-based agents, even if only one of the two players displays social behaviour \linebreak\citep{peysakhovich2017prosocial}. In addition, rewarding actions that lead to a relatively more significant change in the other agent's behaviour may lead to increased cooperation \citep{jaques2019social}. Another study introduces heterogeneity in intrinsic motivation \citep{mckee2020social}. Specifically, the study compares a team of homogeneous agents, who share the same degree of social value orientation, to a heterogeneous group of agents with different degrees of social value orientation. The results show that homogeneous altruistic agents earn relatively high rewards, yet it appears that they adopt a lazy agent approach and produce highly specialised agents. This problem is not evident in heterogeneous groups. Hence, it shows that the widely adopted joint return approach may be undesirable as it masks high levels of inequality amongst agents. 

While studies show that shaping reward functions can lead to better coordination \citep{devlin2011empirical, holmesparker2016combining, peysakhovich2017prosocial, tampuu2017multiagent, jaques2019social, liu2019emergent}, it is very challenging to tune the trade-off between the intrinsic and external reward, and whether it gives rise to cooperative behaviour may depend on the actual task and environment. 

\begin{table*}[!bth]
\centering
\caption{Overview of solutions to the credit assignment problem}
\label{table:creditassignment}
\begin{tabular}{lm{12mm}m{19mm}m{47mm}}
\hline
\multicolumn{1}{c}{\textbf{Study}} &  \multicolumn{1}{c}{\textbf{Implicit/}} & \multicolumn{1}{c}{\textbf{Approach}} & \multicolumn{1}{c}{ \centering{\textbf{Algorithm}}}\\
 & \multicolumn{1}{c}{\textbf{explicit}} &  & \\
\hline
Foerster et al. 2018 & Explicit & Difference rewards            & COMA: uses a counterfactual baseline to marginalise out the action of an agent. \\
Yu et al., 2019 & Explicit & Potential-based rewards            & MA-AIRL: extends maximum entropy inverse reinforcement learning to Markov games. A potential-based function is used to deal with reward shaping ambiguity. \\
Devlin et al., 2014  & Explicit & Difference rewards and potential-based rewards & DriP: uses potential based reward shaping to improve difference rewards. \\
Sunehag et al., 2017 & Implicit & Value-based: deep Q-learning  & VDN: decomposes the team value function into a sum of linear, individual value functions.  \\
Zhou et al., 2020    & Implicit & Policy-based: actor-critic    & LICA: a centralized critic maps current state information into a set of weights, and in turn, mixes individual action vectors into the joint action value estimate.  
\\
\hline
\end{tabular}%
\end{table*}

\subsection{Reward Shaping}

The credit assignment problem refers to the situation when individual agents cannot view their contribution to the joint team reward due to a partially observable environment. Researchers have introduced implicit and explicit reward shaping methods to deal with this problem. Table \ref{table:creditassignment} gives an overview of the reviewed reward shaping methods.

The general solution to this problem is reward shaping, with difference rewards and potential-based reward shaping as the two main classes. Difference rewards consider both the individual and the global reward \citep{foerster2018counterfactual, proper2012modeling, nguyen2018credit, castellini2020difference} and help an agent understand its impact on the environment by removing the noise created by other acting agents. Specifically, it is defined as $D_{i}(z) = G(z) -  G(z - z_{i})$ where $D_{i}$ is the difference reward of agent $i$, $G(z)$ is the global reward considering the joint state-action $z$, and  $G(z - z_{i})$ is a modified version of the state-action vector $z$ in which agent $i$ takes a default action, or more intuitively, the global reward without the contribution of agent $i$ \citep{yliniemi2014multi}. COMA \citep{foerster2018counterfactual} takes inspiration from difference rewards. The centralised critic uses a counterfactual baseline to reason about counterfactuals or alternatives to the state when only that agent's actions change. To marginalise out the action of an agent, an expected value is calculated over all the actions of an agent while keeping other agents' actions constant.
Potential-based reward shaping has also received attention lately \citep{suay2016learning, devlin2014potential}. Formally, it is defined as $F(s, s') = \gamma\Phi(s')-\Phi(s)$ \citep{ng1999policy} where $\Phi(s)$ is a potential function which returns the potential for state $s$ and $\gamma$ is the discount factor. It is a method to incorporate additional information into the reward function to accelerate learning. This approach has been proven not to alter the set of Nash equilibria in a Markov game \citep{devlin2011theoretical}, even when the potential function changes dynamically during learning \citep{devlin2012dynamic}, and combining the two approaches allows agents to converge significantly faster than using difference rewards alone \citep{devlin2014potential}. However, these reward shaping methods require manual tuning for each environment, which is inefficient. Some studies have therefore started looking into the automatic generation of reward shaping, for example, through abstractions derived from an agent's experience \citep{burden2020automating} or via meta-learning on a distribution of tasks \citep{zou2021learning}.

Previous approaches evaluate an agent's action against a baseline to extract its individual effect and belong to the class of explicit credit assignment. In contrast, implicit methods do not work with baselines. Value-based methods decomposes the global value function into individual state-action values, also known as value mixing methods, such as VDN \citep{sunehag2017value}, QMIX \citep{rashid2020monotonic} and QTRAN \citep{son2019qtran} to filter out agent's individual contribution. However, these methods may not handle continuous action spaces effectively. Policy-based algorithms include Learning Implicit Credit Assignment (LICA) \citep{zhou2020learning} and Decomposed multiagent Deep Deterministic Policy Gradient (DE-MADDPG) \citep{sheikh2020multi}. LICA extends the idea of value mixing to policy-based methods. Under the centralised training and decentralised execution framework, a centralised critic is represented by a hypernetwork that maps state information into a set of weights that mixes individual action values into the joint action value. DE-MADDPG extends previous deterministic policy gradient methods using a dual-critic framework. The global critic takes as input all agents' observations and actions and estimates the global reward. The local critic receives as input only the local observation and action of an agent and estimates the local reward. This framework achieves better and more stable performance than earlier deterministic policy gradient methods. 

\section{Discussion}
\label{sec:discussion}

We have surveyed a range of studies in DMARL. While integrating deep neural networks in reinforcement learning has dramatically improved agents' learning in more complex and larger environments, we wish to highlight current limitations and open challenges in the field.

\begin{itemize}
\item In the development from single-agent reinforcement learning to multiagent reinforcement learning, most earlier studies used a game-theoretic lens to study interactive decision-making, assuming perfectly rational agents who maximise their behaviour through a deliberate optimisation process. However, while game theory's strength lies in its generalizability and mathematical precision, experiments have shown that it is often a poor representation of actual human behaviour \citep{colman2003cooperation}. Researchers must consider irrational and altruistic decision-making, especially if we wish to extend artificial intelligence (AI) to more realistic environments or design applications for human-AI interaction in larger and more complex problems. We have seen that pro-social agents can achieve better group outcomes \citep{peysakhovich2017prosocial, hughes2018inequity}. However, studies are still limited, and we encourage fellow researchers to deepen our understanding in this field.  

\item We also want to bring attention to the design and assumptions in current research. Many studies assume homogeneous agents; from a practical viewpoint, this may accelerate learning since agents can share policies and parameters. Agents thus only need to learn one policy and may better anticipate the behaviour of other agents. However, whether this also leads to better performance in the final task is an open question. For instance, a soccer team usually consists of forwards, midfielders, defenders and a goalkeeper. The team's success is partly determined by how well each fulfils these different roles. Thus, an interesting question is whether letting each agent learn its own policy and have heterogeneous teams pays off. While homogeneous agents can still act differently due to different observations input, the observation space must be the same size. This assumption does not always hold. For instance, agents have different observation spaces in soccer as individuals occupy different positions on the field. Preliminary results show that despite making the learning slower at the beginning, heterogeneous teams perform better at the final task \citep{kurek2016heterogeneous}. Another study provides formal proof for parameter sharing between heterogeneous agents \citep{terry2021revisiting}, which may mitigate the slow start problem. 

\item Studies may also rely on unrealistic assumptions. For instance, multiple studies require access to opponents' information, such as trajectories or parameters, while their problem domain actually gives an incentive to hide information. Others assume fixed behaviours of agents or that agents can view the global state. 

\item Another issue is the generalizability of studies. For example, many studies require handcrafted features or rewards specific to the environment. In addition, a majority of the studies are evaluated in two-player games. As a result, a danger exists that the agent's policy overfits to the behaviour of the second agent (i.e. the lazy agent problem) and does not generalise to other settings. Future research should integrate more realistic assumptions and work on the generalizability of studies to settings with more players or different environments.

\end{itemize}

While DMARL has seen a significant improvement in the types and complexities of challenges it can address, several hurdles remain. For example, problems associated with large search spaces, partially observable environments, nonstationarity, sparse rewards and the exploration-exploitation trade-off remain challenging. These issues are partly due to computational constraints, such that assumptions are often relaxed. We want to point out two other research areas, namely evolutionary algorithms and psychology, that may help researchers address some of the open questions. 

\subsection{Evolutionary Algorithms}

Evolutionary algorithms (EAs) are inspired by nature's creativity and simulate the process of organic evolution to solve optimisation problems. 
In simple terms, a randomly initialised population of individual solutions evolves toward better regions of the search space via selection, mutation and recombination operators. 
A fitness function evaluates the quality of the individuals and favours the reproduction of those with a higher fitness score, while mutation maintains diversity in the population \citep{back1993overview}. 
An early study sheds light on how EAs  deal with RL problems \citep{moriarty1999evolutionary} and has been confirmed by recent studies \citep{bloembergen2015evolutionary,drugan2019reinforcement, arulkumaran2019alphastar,lehman2018es, lehman2018safe,conti2018improving, such2018deep,zhang2017relationship}. 
EAs offer a novel perspective to scaling RL multiagent systems as it is highly parallelisable, and there is no need for backpropagation \citep{such2018deep,majumdar2020evolutionary}. 

EAs have been compared with popular value-based and policy-gradient algorithms such as DQN and A3C \citep{such2018deep}. 
Novelty search \citep{such2018deep, conti2018improving} is a promising area \citep{lehman2008exploiting} 
since it encourages exploration on tasks with sparse rewards and deceptive local optima---problems that remain an issue with conventional reward-maximising methods. 
EAs have been shown to work well with nonstationarity and partial observability, as it continually uses and evolves a population of agents instead of a single agent \citep{moriarty1999evolutionary, liu2020mapper}. EAs can evolve agents with different policies \citep{gomes2017dynamic,gomes2014avoiding,nitschke2012evolving}, such that heterogeneity can be introduced in team-based learning. 
Population-based training has proven powerful in achieving superhuman behaviour in Capture the Flag \citep{jaderberg2019human}, and StarCraft \citep{vinyals2019grandmaster}. 

\subsection{Psychology}
Many key ideas in reinforcement learning, such as operant conditioning and trial-and-error, originated in psychology and cognitive science research
\citep{sutton1998introduction}. Interestingly, recent DMARL studies started moving towards more human-like agents, showing that characteristics like reciprocity and intrinsic motivation pay off. 

We believe psychology may provide more valuable insights into current problems in DMARL. For instance, bounded rationality models \citep{simon1990bounded, simon1957models} describe how individuals make decisions under a finite amount of knowledge, time and attention. To deal with bounded rationality, people use heuristics, or mental shortcuts, to solve problems quickly and efficiently \citep{gigerenzer1996reasoning}. While RL research already uses heuristics to deal with large and complex problems \citep{cheng2021heuristic, ma2021distributed}, selecting suitable heuristics is still insufficiently explored. Psychology has a long tradition of investigating heuristics and may offer new perspectives. In addition,  heuristics aid in filtering relevant information in a complex world, which may benefit agents in partially observable environments or counter negative knowledge transfer \citep{marewski2010good}. However, intuitive judgement can also lead to biases and suboptimal decision-making \citep{gilovich2002heuristics}.

Humans are also capable of creative problem-solving, a prerequisite for innovation. Likewise, agents need to explore the environment to find more optimal solutions. A first approach of combining creativity with reinforcement learning shows that creativity offers the potential to explore promising solution spaces, whereas traditional methods fail  \citep{colin2016hierarchical}.

Lastly, psychology can play an essential role in helping researchers understand how agents make decisions and tackle the black-box problem of deep neural networks. Cognitive psychologists have developed robust models of human behaviour, such as decision making, attention and language, without observing these processes directly but through controlled behavioural experiments in which cognitive functions can be isolated \citep{taylor2021artificial}. Open-source platforms are now also available \citep{leibo2018psychlab} that allow researchers to use methods from cognitive psychology to study the behaviours of artificial agents in a controlled environment. We encourage researchers to draw from psychology research and its methodologies to analyse agents' complex interactions and better understand and improve their decision-making.

\section{Conclusion}
\label{sec:conclusion}

The current survey has presented an overview of the challenges inherent in multiagent representations. We have identified five different research areas in DMARL that aim to mitigate one or multiple of these challenges: (1) centralised training and decentralised execution, (2) opponent modelling, (3) communication, (4) efficient coordination, and (5) reward shaping. While early studies drew inspiration from game theory and were evaluated on grid-based games, the field is moving towards more sophisticated and realistic representations. Nevertheless, dealing with large problem spaces and sparse rewards in nonstationary and partially observable settings remains an open issue. 

Existing research has approached this problem mainly from traditional, computational, RL perspectives. While combining deep learning with value-based and policy-based methods has been shown to mitigate the problem, they seem to be only part of the answer. We encourage researchers to take an interdisciplinary perspective on developing new solutions and benefit from the knowledge of other research domains. Specifically, evolutionary algorithms offer insights into dealing with larger, nonstationary and partially observable environments. At the same time, sociology and psychology increase our understanding of agents' reasoning patterns and offer us alternatives in dealing with sparse rewards, such as intrinsic motivation. Finally, we believe that integrating multiple research disciplines leads to more realistic scenarios humans encounter in practice, so the findings may eventually be fruitful in real-world applications.


%
%
\section{Declarations}
\subsection{Funding}
No funding was received to assist with the preparation of this manuscript.
\subsection{Conflicts of interests}
The authors have no conflicts of interest to declare that are relevant to the content of this article.
\subsection{Availability of data and material}
Not applicable
\subsection{Code availability}
Not applicable
\subsection{Data availability}
Data sharing not applicable to this article as no datasets were generated or analysed during the current study.

\bibliographystyle{spbasic}      
\bibliography{references}   

\end{document}